\documentclass[sigconf, 10pt, anonymous=false]{acmart}

\renewcommand\footnotetextcopyrightpermission[1]{} 
\newcommand\blfootnote[1]{%
  \begingroup
  \renewcommand\thefootnote{}\footnote{#1}%
  \addtocounter{footnote}{-1}%
  \endgroup
}

\AtBeginDocument{%
  \providecommand\BibTeX{{%
    \normalfont B\kern-0.5em{\scshape i\kern-0.25em b}\kern-0.8em\TeX}}}

\acmYear{2020}\copyrightyear{2020}
\setcopyright{rightsretained}
\acmConference[Accepted toEMDL '20]{4th International Workshop on Embedded and Mobile Deep Learning}{September 21, 2020}{London, United Kingdom}
\acmBooktitle{4th International Workshop on Embedded and Mobile Deep Learning (EMDL '20), September 21, 2020, London, United Kingdom}
\acmPrice{}
\acmDOI{10.1145/3410338.3412338}
\acmISBN{978-1-4503-8073-7/20/09}

\usepackage{booktabs}
\usepackage{colortbl}
\usepackage{multirow}
\usepackage{subcaption}
\usepackage{tablefootnote}
\usepackage{tabularx}
\usepackage{url}
\usepackage{natbib}

\newcommand{\RNum}[1]{\uppercase\expandafter{\romannumeral #1\relax}}

\begin{document}

\title{Split Computing for Complex Object Detectors: Challenges and Preliminary Results}

\author{Yoshitomo Matsubara}
\affiliation{\institution{University of California, Irvine}}
\email{yoshitom@uci.edu}

\author{Marco Levorato}
\affiliation{\institution{University of California, Irvine}}
\email{levorato@uci.edu}


\renewcommand{\shortauthors}{Matsubara and Levorato}

\begin{abstract}
Following the trends of mobile and edge computing for DNN models, an intermediate option, split computing, has been attracting attentions from the research community.
Previous studies empirically showed that while mobile and edge computing often would be the best options in terms of total inference time, there are some scenarios where split computing methods can achieve shorter inference time.
All the proposed split computing approaches, however, focus on image classification tasks, and most are assessed with small datasets that are far from the practical scenarios.
In this paper, we discuss the challenges in developing split computing methods for powerful R-CNN object detectors trained on a large dataset, COCO 2017.
We extensively analyze the object detectors in terms of layer-wise tensor size and model size, and show that naive split computing methods would not reduce inference time.
To the best of our knowledge, this is the first study to inject small bottlenecks to such object detectors and unveil the potential of a split computing approach.
The source code and trained models' weights used in this study are available at \textbf{\url{https://github.com/yoshitomo-matsubara/hnd-ghnd-object-detectors}}.
\end{abstract}





\maketitle

\pagestyle{empty} 
\blfootnote{Accepted to 4th International Workshop on Embedded and Mobile Deep Learning (EMDL '20), September 21, 2020, London, United Kingdom}

\section{Introduction}
\label{sec:intro}
Along with the rapid evolution of computing devices, deep learning approaches have been widely studied to train powerful machine learning models, and many of such models achieve state-of-the-art performance in various tasks such as vision and natural language processing.
Such models, however, are often too complex to be executed on mobile devices, which have a severely constrained computational capacity.
To address this critical problem, the research community proposed two key strategies: model compression and edge computing.
In the former approach, complex models are simplified, \emph{e.g.}, by knowledge distillation~\cite{Hinton14} and model pruning and quantization~\cite{polino2018model,wei2018quantization}.
In the latter approach~\cite{Barbera13,Mach17,Liu19}, mobile devices offload the execution of computationally expensive models to powerful (edge) computers located at the network edge.
Clearly, edge computing requires to wirelessly transport the input data and model outcome on the link connecting the mobile device to the edge computer.

Recently, an intermediate option, namely \emph{split Deep Neural Network} (DNN) or \emph{split computing}, has been attracting a considerable interest~\cite{Kang17,Teerapittayanon17,Li18,eshratifar2019bottlenet,Matsubara19,Emmons19,hu2020fast,shao2020bottlenet++}.
Many of such methods literally split DNN models into head and tail portions, which are executed by the mobile device and edge computer, respectively. Note that in this case, instead of the input data, the tensor produced by the head model should be transported to the edge computer.
More sophisticated approaches~\cite{eshratifar2019bottlenet,Matsubara19,hu2020fast,shao2020bottlenet++} modify the architecture of the model itself to (a) reduce the size of the tensor to be transferred, and (b) reduce the computing load assigned to the mobile device. Intuitively, the two features mentioned above would prove essential in enabling effective task offloading in challenged settings (\emph{e.g.}, low capacity of the wireless channel and low computing capacity at the mobile device).

Although promising, as discussed in detail in Section~\ref{subsec:split_comp}, most of the existing split DNN approaches are either not evaluated~\cite{Emmons19} or evaluated only in simple classification tasks~\cite{Teerapittayanon17,eshratifar2019bottlenet,Matsubara19,hu2020fast,shao2020bottlenet++}, such as on miniImageNet, Caltech 101, CIFAR -10, and -100 datasets.
The goal of this paper is to discuss the several technical challenges in achieving effective DNN splitting for edge computing in one of the most difficult computing tasks, that is, object detection.
We specifically consider Faster and Mask R-CNNs~\cite{Ren15,He17}.

Our extensive module-wise model analysis indicates that naive splitting strategies would fail to provide any improvements, where performance is measured in total inference time, compared to mobile and edge computing.
Then, we propose to redesign the object detectors to introduce small bottlenecks whose output tensor sizes are significantly smaller than that of the input layer.
To the best of our knowledge, this is the first work that discusses split DNN approaches for such powerful object detectors providing 1) benchmark results on a well-known object detection dataset, COCO 2017, and 2) a thorough illustration of the tradeoff between bottleneck tensor size and detection performance.

As discussed in detail in Section~\ref{subsec:split_comp}, the complex structure of CNN-based object detectors poses unique challenges in designing effective splitting approaches.
For instance, as the models branch to provide intermediate outputs to later modules, splitting in later layers would require to send multiple tensors, which results in an increased amount of data to be transferred.
Our design, then, places the bottleneck early in the model, before the first branching.
However, this strategy requires an effective re-design of the network, as the first layers tend to amplify the input, rather than compressing it.
The bottlenecks we designed reduced the tensor size by about 80.3 -- 93.4$\%$ compared to pure edge computing (where the input tensor is transmitted), and the aggressively small bottlenecks resulted in significant loss of the detection performance.
Based on our study, it is apparent that there is a strong need for efficient training and compression strategies (\emph{e.g.}, quantization) to achieve effective splitting in a wide range of settings and conditions.
The source code and trained models' weights used in this study are released~\footnote{\label{fn:our_repo}\url{https://github.com/yoshitomo-matsubara/hnd-ghnd-object-detectors}} to enable such further studies and developments.

\section{CNN-based Object Detectors}
\label{sec:models}
In this section, we discuss the architecture of recent object detectors based on Convolutional Neural Network (CNN) that achieve state-of-the-art detection performance.
These CNN-based object detectors are often categorized into either single-stage or two-stage models.
Single-stage models, such as YOLO and SSD series~\cite{Redmon16,Liu16}, are designed to directly predict bounding boxes and classify the contained objects. Conversely, two-stage models~\cite{Ren15,He17} generate region proposals as output of the first stage, and classify the objects in the proposed regions in the second stage.
In general, single-stage models have smaller execution time due to their lower overall complexity compared to two-stage models, that are superior to single-stage ones in terms of detection performance.

\begin{figure}[t]
    \centering
    \includegraphics[width=0.59\linewidth]{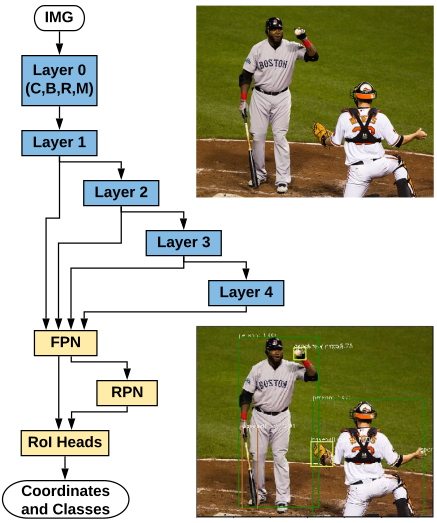}
    \vspace{-1em}
    \caption{R-CNN with ResNet-based backbone. Blue modules are from its backbone model, and yellow modules are of object detection. C: Convolution, B: Batch normalization, R: ReLU, M: Max pooling layers.}
    \label{fig:rcnn_detector}
\end{figure}

Recent object detectors, \emph{e.g.}, Mask R-CNN and SSD~\cite{He17,Sandler18}, adopt state-of-the-art image classification models, such as ResNet~\cite{He16} and MobileNet v2~\cite{Sandler18}, as \emph{backbone}.
The main role of backbones in detection models pretrained on large image datasets, such as the ILSVRC dataset, is feature extraction.
As illustrated in Figure~\ref{fig:rcnn_detector}, such features include the outputs of multiple intermediate layers in the backbone.
All the features are fed to complex modules specifically designed for detection tasks, \emph{e.g.}, the feature pyramid network~\cite{Lin17}, to extract further high-level semantic feature maps at different scales.
Finally, these features are used for bounding box regression and object class prediction.

In this study, we focus our attention on state-of-the-art two-stage models.
Specifically, we consider Faster R-CNN and Mask R-CNN~\cite{Ren15,He17} pretrained on the COCO 2017 datasets.
Faster R-CNN is the strong basis of several 1st-place entries~\cite{He16} in ILSVRC and COCO 2015 competitions. The model is extended to Mask R-CNN by adding a branch for predicting an object mask in parallel with the existing branch for bounding box recognition~\cite{He17}.
Mask R-CNN not only is a strong benchmark, but also a well-designed framework, as it easily generalizes to other tasks such as instance segmentation and person keypoint detection.


\section{Challenges and Approaches}
\label{sec:challenges}
We discuss challenges in deploying CNN-based object detectors in three different scenarios: mobile, edge, and split computing.
We use total inference time (including the time needed to transport the data over the wireless channel) and object detection performance as performance metrics.

\subsection{Mobile and Edge Computing}
In mobile computing the mobile device executes the whole model, and the inference time is determined by the complexity of the model and local computing power. Due to limitations in the latter, in order to have tolerable inference time, the models must be simple and lightweight.
To this aim, one of the main approaches is to use human-engineered features instead of those extracted from stacked neural modules.
For instance, Mekonnen \emph{et al.}~\cite{mekonnen2013fast} propose an efficient HOG (Histogram Oriented Gradients) based person detection method for mobile robotics.
Designing high-level features of human's behavior on touch screen, Matsubara \emph{et al.}~\cite{matsubara2016screen} propose distance/SVM-based one-class classification approaches to screen unlocking on smart devices in place of password or fingerprint authentications.

In recent years, however, deep learning based methods have been outperforming the models with human-engineered features in terms of accuracy.
For image classification tasks, MobileNets~\cite{Sandler18,howard2019searching} and MNasNets~\cite{tan2019mnasnet} are examples of models designed to be executed on mobile devices, while providing moderate classification accuracy.
Corresponding lightweight object detection models are SSD~\cite{Liu16} and SSDLite~\cite{Sandler18}.
Techniques such as model pruning, quantization and knowledge distillation~\cite{polino2018model,wei2018quantization,Hinton14} can be used to produce lightweight models from larger ones.

Table~\ref{table:tf_map_coco2014} summarizes the performance of some models trained on the  COCO 2014 minival dataset as reported in the TensorFlow repository~\footnote{\label{fn:tf_repo}\url{https://github.com/tensorflow/models/blob/master/research/object_detection/g3doc/detection_model_zoo.md\#coco-trained-models}}.
Obviously, the SSD series object detectors with MobileNet backbones outperform the Faster R-CNN with ResNet-50 backbone in terms of inference time, but such lightweight models offer degraded detection performance. 
Note that in the repository, the COCO 2014 minival split is used for evaluation, and the inference time is measured on a machine with one NVIDIA GeForce GTX TITAN X, which is clearly not suitable to be embedded in a mobile device.
Also, the values reported in Table~\ref{table:tf_map_coco2014} are given for models implemented with the TensorFlow framework with input images resized to 600 $\times$ 600, though some of the models in the original work such as Faster R-CNN~\cite{Ren15} use different resolutions.
In general, the classification/detection performance is often compromised in mobile computing due to their limited computing power.

\begin{table}[t]
\caption{Mean average precision (mAP) on COCO 2014 minival dataset and running time on a machine with an NVIDIA GeForce GTX TITAN X.~\textsuperscript{\ref{fn:tf_repo}}}
\vspace{-1em}
\begin{center}
    \begin{tabular}{l|rr} \toprule
    \multicolumn{1}{c|}{TensorFlow model} & \multicolumn{1}{c}{mAP} & \multicolumn{1}{c}{Speed [sec]} \\ \midrule
    SSDLite with MobileNet v2 & 0.220 & 0.027 \\ 
    SSD with MobileNet v3 (Large) & 0.226 & N/A* \\ 
    Faster R-CNN with ResNet-50 & 0.300 & 0.0890 \\ \bottomrule
	\end{tabular}
\end{center}
\small
\raggedright * Reported speed was measured on a different device
\label{table:tf_map_coco2014}
\vspace{0.5em}


\caption{Running time [sec/image] of Faster and Mask R-CNNs with different ResNet models.}
\vspace{-1em}
\begin{center}
\bgroup
\setlength{\tabcolsep}{0.5em}
\def\arraystretch{1.2}
    \begin{tabular}{cl|rrrr} \toprule
    \multicolumn{2}{c|}{Backbone with ResNets} & \multicolumn{1}{c}{-18} & \multicolumn{1}{c}{-34} & \multicolumn{1}{c}{-50} & \multicolumn{1}{c}{-101} \\ \midrule
    \multirow{3}{*}{\rotatebox{90}{\scriptsize Faster R-CNN}} & Raspberry Pi 4 Model B & 27.73 & 23.40 & 26.14 & 35.16 \\ 
    & NVIDIA Jetson TX2 & 0.617 & 0.743 & 0.958 & 1.26 \\ 
    & Desktop + 1 GPU & 0.0274 & 0.033 & 0.0434 & 0.0600 \\ \midrule
    \multirow{3}{*}{\rotatebox{90}{\scriptsize Mask R-CNN}} & Raspberry Pi 4 Model B & 18.30 & 23.65 & 27.02 & 34.73 \\
    & NVIDIA Jetson TX2 & 0.645 & 0.784 & 0.956 & 1.27 \\
    & Desktop + 1 GPU & 0.0289 & 0.0541 & 0.0613 & 0.0606 \\\bottomrule
	\end{tabular}
\egroup
\end{center}
\label{table:rcnns_time}
\end{table}

Table~\ref{table:rcnns_time} highlights that it would be impractical to deploy some of the powerful object detectors on weak devices.
Specifically, on Raspberry Pi 4, R-CNN object detectors with even the smallest backbone in the family took 20--30 seconds for prediction per a resized image of which shorter side resolution is 800 pixels, following~\cite{Ren15,He17}.
For both Faster and Mask R-CNNs, the execution time on NVIDIA Jetson TX2 and a desktop machine with an NVIDIA RTX 2080 Ti is sufficiently small to support real-time applications .

Different from mobile computing, the total inference time in edge computing is the sum of the execution time in Table~\ref{table:rcnns_time} and the communication time needed to transfer data from the mobile device to the edge computer (\emph{e.g.}, Raspberry Pi 4 and the desktop machine, respectively).
If the prediction results are to be sent back to the mobile device, a further communication delay term should be taken into account, although outcomes (\emph{e.g.}, bounding boxes and labels) typically have a much smaller size compared to the input image.
As discussed in~\cite{Kang17,Matsubara19}, the delay of the communication from mobile device to edge computer is a critical component of the total inference time, which may become dominant in some network conditions, where the performance of edge computing may suffer from a reduced channel capacity.



\subsection{Split Computing}
\label{subsec:split_comp}

Split computing is an intermediate option between mobile and edge computing. The core idea is to split models into head and tail portions, which are deployed at the mobile device and edge computer, respectively.
To the best of our knowledge, Kang \emph{et al.}~\cite{Kang17} were the first to propose to split deep neural network models. However, the study simply proposed to optimize where to split the model, leaving the architecture unaltered.

In split computing, the total inference time is sum of three components: mobile processing time, communication delay, and edge processing time.
To shorten the inference time in split computing compared to those of mobile and edge computing, the core challenge is to significantly reduce communication delay while leaving a small portion of computational load on mobile device for compressing the data to be transferred to edge server.
Splitting models in a straightforward way, as suggested in~\cite{Kang17}, however, does not lead to an improvement in performance in most cases.
The tension is between the penalty incurred by assigning a portion of the overall model to a weaker device (compared to the edge computer) and the potential benefit of transmitting a smaller amount of data.
However, most models do not present ``natural'' bottlenecks in their design, that is, layers with a small number of output nodes, corresponding to a small tensor to be propagated to the edge computer.
In fact, the \emph{neurosurgeon} framework locates pure mobile or edge computing as the optimal computing strategies in most models.

Building on the work of Kang \emph{et al.}~\cite{Kang17}, recent contributions propose DNN splitting methods~\cite{Teerapittayanon17,Li18,eshratifar2019bottlenet,Matsubara19,Emmons19,hu2020fast,shao2020bottlenet++}.
Most of these studies, however, (\RNum{1}) do not evaluate models using their proposed lossy compression techniques~\cite{Emmons19}, (\RNum{2}) lack of motivation to split the models as the size of the input data is exceedingly small, \emph{e.g.}, 32 $\times$ 32 pixels RGB images in~\cite{Teerapittayanon17,hu2020fast,shao2020bottlenet++}, (\RNum{3}) specifically select models and network conditions in which their proposed method is advantageous~\cite{Li18}, and/or (\RNum{4}) assess proposed models in simple classification tasks such as miniImageNet, Caltech 101, CIFAR -10, and -100 datasets~\cite{eshratifar2019bottlenet,Matsubara19,hu2020fast,shao2020bottlenet++}.

Similar to CNN-based image classification models, it is not possible to reduce the inference time of CNN-based object detectors by naive splitting methods without altering the models' architecture.
This is due to the designs of the early layers of the models, which \emph{amplify} the input data size.
It would be worth noting that \citet{Matsubara19} apply a loseless compression technique, a standard Zip compression, to intermediate outputs of all the splittable layers in a CNN model, and show the compression gain is not sufficient to significantly reduce inference time in split computing.

\begin{figure}[t]
    \centering
    \includegraphics[width=0.85\linewidth]{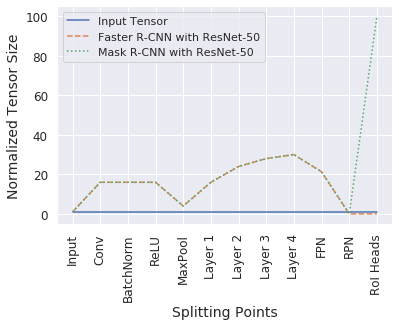}
    \vspace{-1.5em}
    \caption{Layer-wise output tensor sizes of Faster and Mask R-CNNs scaled by input tensor size ($3 \times 800 \times 800$).}
    \label{fig:rcnns_tensor_sizes}
\end{figure}

Figure~\ref{fig:rcnns_tensor_sizes} illustrates this effect by showing the amplification of the data at each of core layers in Faster and Mask R-CNNs with ResNet-50, compared to the input tensor size ($3 \times 800 \times 800$).
Note that these models are designed for images whose shorter side resolution is 800 pixels~\cite{Ren15,He17}.
The trends confirm that there is no splitting point (below blue line) in any of the early layers.
Therefore, naive splitting does not result in any gain in terms of communication delay.
We note that different from the Faster R-CNN model, the output tensor of the RoI Heads in the Mask R-CNN model is significantly larger than the input tensor.
As the model emits not only bounding boxes and object classes, but also pixel-level masks for segmentation, the last tensor size surges in Figure~\ref{fig:rcnns_tensor_sizes} (green dotted line), but the general trend looks the same with Faster R-CNN model when using bounding boxes and object classes only.

    

A promising, but challenging, solution to reduce the inference time in challenged networks is to introduce a small \emph{bottleneck} within the model, and split the model at that layer~\cite{Matsubara19}.
In the following section, we discuss bottleneck injection for CNN-based object detectors, specifically Faster and Mask R-CNNs, and present preliminary experimental results supporting this strategy.

\section{Experiments}
One of the core challenges of bottleneck injection in R-CNN object detectors is that the bottleneck needs to be introduced in early stages of the detector compared to image classification models.
As illustrated in Figure~\ref{fig:rcnn_detector}, the first \emph{branch} of the network is after Layer 1.
As a result, the bottleneck needs to be injected before the layer to avoid the need to forward multiple tensors produced by the branches (Figs.~\ref{fig:rcnn_detector} and \ref{fig:rcnns_tensor_sizes}).

\begin{figure}[t]
    \centering
    \includegraphics[width=0.85\linewidth]{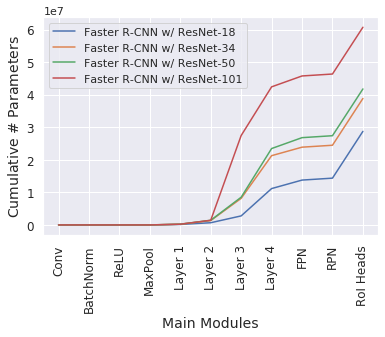}
    \vspace{-1.5em}
    \caption{Cumulative number of parameters in core modules of \underline{Faster R-CNN}.}
    \label{fig:faster_rcnn_params}
    
    \includegraphics[width=0.85\linewidth]{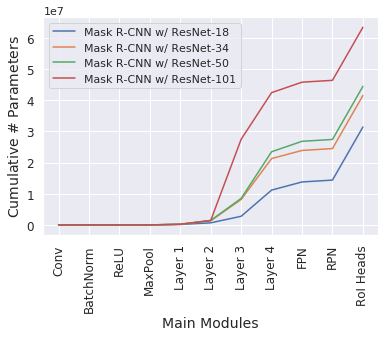}
    \vspace{-1.5em}
    \caption{Cumulative number of parameters in core modules of \underline{Mask R-CNN}.}
    \label{fig:mask_rcnn_params}
\end{figure}

The amount of computational load assigned to the mobile device should be considered as well when determining the bottleneck placement. In fact, the execution time of the \emph{head} model, which will be deployed on the mobile device, is a critical component to minimize the total inference time.
Figures~\ref{fig:faster_rcnn_params} and \ref{fig:mask_rcnn_params} depict the number of parameters of each model used for partial inference on the mobile device when splitting the model at specific modules. The reported values provide a rough estimate of the head model's complexity as a function of the splitting point.

Recall that feature pyramid network (FPN), region proposal network (RPN), and region of interest (RoI) Heads in the R-CNN models are designed specifically for object detection tasks, and all the modules before them are originally from an image classification model (ResNet models~\cite{He16} in this study).
Because of not only the models' branching, but the trends in these figures, it is clear that the bottleneck, and thus the splitting point, should be placed before ``Layer 1''.

Matsubara \emph{et al.}~\cite{Matsubara19} attempted to introduce a bottleneck in the first convolution layer of DenseNet-169~\cite{Huang17}.
The bottleneck uses 4 output channels in place of 64 channels, so that the output tensor of the layer is smaller than the input tensor to the model.
Using Caltech 101 dataset, they naively trained the redesigned model, that significantly degraded classification accuracy even despite the relative low complexity of the dataset compared to the ILSVRC dataset.

Based on the analysis and results, we attempt to introduce a bottleneck to ``Layer 1'', that consists of multiple low-level modules such as convolution layers.
Here, we redesign the layer 1 by pruning a couple of layers and adjust hyperparameters to make its output shape match that of the layer 1 in the original model.
The redesigned layer 1 has a small bottleneck with $C$ output channels, a key parameter to control the balance between detection performance and bottleneck size.

Assuming that the original models are overparameterized, we \emph{distill} the head models while injecting bottlenecks.
Specifically, we use head network distillation~\cite{Matsubara19}, a teacher-student training scheme, only applied to the head portion to reduce training time.
We treat the first layers until the layer 1 in the original detector as a teacher model, and those in the redesigned detector as a student model.
Note that the redesigned detector reuses all the modules and learnt parameters of the original detector except the modules until the end of the layer 1.
The exact network architectures are not described in this paper due to limited space, but the code and trained model weights are released to ensure reproducible results.~\textsuperscript{\ref{fn:our_repo}}
In this study, we use pretrained Faster and Mask R-CNNs with ResNet-50 as teacher models.
To the best of our knowledge, this is the first study that discusses introducing bottlenecks to CNN-based object detectors for split computing and provide experimental results.

\begin{figure}[t]
    \centering
    \includegraphics[width=0.85\linewidth]{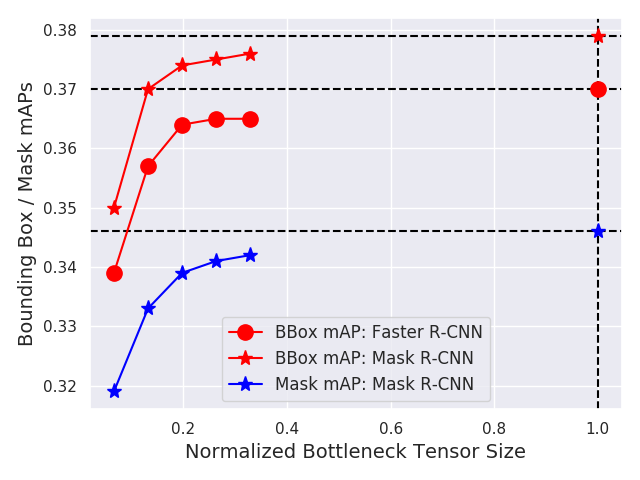}
    \vspace{-1.5em}
    \caption{Average bottleneck tensor size vs. BBox and Mask mAPs on COCO 2017 validation dataset (test split is not publicly available). Bottleneck tensor size is scaled by average input tensor size ($3 \times 874 \times 1044$).}
    \label{fig:bottleneck_vs_map}
\end{figure}

Figure~\ref{fig:bottleneck_vs_map} illustrates the relationship between the output tensor size of the introduced bottleneck with the number of output channels $C = 3, 6, 9, 12, 15$ and bounding box / mask mAPs of the modified Faster and Mask R-CNNs.
The rightmost data points on the dashed line in the figure correspond to the detection performance (mAP) of the original R-CNN models in edge computing, that is, the splitting points are at their input layers.
Since in this study we do not alter the tail portion of the models, but modify and train the head portion only, we take the detection performance of the original models (on dashed line) as the upper bound performance of our modified models. It can be observed how we successfully reduce the size of the tensors to be transferred to the edge computer compared to the input tensors incurring some mAP degradation.
Recall that there are no effective splitting points in the original models, as shown in Figure~\ref{fig:rcnns_tensor_sizes} (\emph{i.e.}, most of the normalized tensor sizes are above 1), and our introduced bottlenecks save approximately 80.3 -- 93.4$\%$ of tensor size for offloading compared to edge computing.


\section{Conclusions and Future Work}
\label{sec:conclusions}
In this study, we discussed the challenges in deploying CNN-based object detectors in mobile devices using three key strategies: mobile computing, edge computing, and the recently proposed split computing.
We focused our discussion on two different state-of-the-art two-stage object detectors, which have no suitable ``natural'' splitting points, and injected small bottlenecks into the models based on the analysis we provided.
While the introduced bottlenecks are smaller than the input tensors, the detection performance is sometimes significantly degraded specifically when introducing aggressively small bottlenecks.
In addition to improving the detection performance, it would be necessary to assess the inference time using further compression techniques such as quantization, which we further discuss in~\cite{matsubara21neural}.



\begin{acks}
This work was supported by the NSF grant IIS-1724331 and MLWiNS-2003237, and DARPA grant HR00111910001.
\end{acks}

\bibliographystyle{ACM-Reference-Format}
\bibliography{references}










\end{document}